\documentclass[sigconf,authorversion,nonacm]{acmart}
\AtBeginDocument{%
  \providecommand\BibTeX{{%
    \normalfont B\kern-0.5em{\scshape i\kern-0.25em b}\kern-0.8em\TeX}}}

\usepackage{color-edits}
\usepackage{caption}
\usepackage{subfigure}
\usepackage[capitalize,noabbrev]{cleveref}
\usepackage{enumitem}
\usepackage{tikz}

\usepackage[flushleft]{threeparttable}
\AtBeginEnvironment{tablenotes}{\small}
\addauthor{yk}{red}
\addauthor{ch}{blue}

\begin{document}

\title{Leveraging Large Language Models for Hybrid Workplace Decision Support}\thanks{Some of the information in this document relates to pre-released content which may be subsequently modified. Microsoft makes no warranties, express or implied, with respect to the information provided here.  This document is provided ``as-is". Information and views expressed in this document, including URL and other Internet Web site references, may change without notice.  Some examples depicted herein are provided for illustration only and are fictitious. No real association or connection is intended or should be inferred.  This document does not provide you with any legal rights to any intellectual property in any Microsoft product.

© 2024 Microsoft. All rights reserved.}

\author{Yujin Kim}
\email{kimyujin@microsoft.com}
\affiliation{%
  \institution{Office of Applied Research, Microsoft}
  \streetaddress{One Microsoft Way}
  \city{Redmond}
  \state{Washington}
  \country{USA}
  \postcode{98052}
}

\author{Chin-Chia Hsu}
\email{chinchiahsu@microsoft.com}
\affiliation{%
  \institution{Office of Applied Research, Microsoft}
   \streetaddress{One Microsoft Way}
  \city{Redmond}
  \state{Washington}
  \country{USA}
  \postcode{98052}
  }

\renewcommand{\shortauthors}{Kim and Hsu}

\begin{abstract}
  Large Language Models (LLMs) hold the potential to perform a variety of text processing tasks and provide textual explanations for proposed actions or decisions. In the era of hybrid work, LLMs can provide intelligent decision support for workers who are designing their hybrid work plans. In particular, they can offer suggestions and explanations to workers balancing numerous decision factors, thereby enhancing their work experience. In this paper, we present a decision support model for workspaces in hybrid work environments, leveraging the reasoning skill of LLMs. We first examine LLM's capability of making suitable workspace suggestions. We find that its reasoning extends beyond the guidelines in the prompt and the LLM can manage the trade-off among the available resources in the workspaces. We conduct an extensive user study to understand workers' decision process for workspace choices and evaluate the effectiveness of the system. We observe that a worker's decision could be influenced by the LLM's suggestions and explanations. The participants in our study find the system to be convenient, regardless of whether reasons are provided or not. Our results show that employees can benefit from the LLM-empowered system for their workspace selection in hybrid workplace. 

\end{abstract}


\begin{CCSXML}
<ccs2012>
   <concept>
       <concept_id>10002951.10003227.10003241</concept_id>
       <concept_desc>Information systems~Decision support systems</concept_desc>
       <concept_significance>500</concept_significance>
       </concept>
   <concept>
       <concept_id>10003120.10003121.10003122.10003334</concept_id>
       <concept_desc>Human-centered computing~User studies</concept_desc>
       <concept_significance>500</concept_significance>
       </concept>
    <concept>
       <concept_id>10010147.10010178.10010179</concept_id>
       <concept_desc>Computing methodologies~Natural language processing</concept_desc>
       <concept_significance>500</concept_significance>
       </concept>
 </ccs2012>
\end{CCSXML}

\ccsdesc[500]{Information systems~Decision support systems}
\ccsdesc[500]{Human-centered computing~User studies}
\ccsdesc[500]{Computing methodologies~Natural language processing}

\keywords{Large language model, decision support system, hybrid workplace, explainable artificial intelligence}


\received{25 January 2024}

\maketitle

\section{Introduction}
The COVID-19 pandemic sparked a significant transition from office-based work to remote work in 2020.
This global-scale impact on humans' daily activities prompted both organizations and workers to rethink the future of work, moving beyond the traditional five-day office model \cite{teevan2022microsoft}.
Indeed, in the  post-pandemic era, numerous companies are adopting the hybrid work model, offering their employees the flexibility to decide when and where they work \cite{barrer2021working}.
It is no longer necessary for employees to be physically present in offices every workday.

In hybrid work environments, many companies are motivated to adopt a desk sharing (or hot desking) approach \cite{Brue2023New}, in which multiple employees share a single desk based on their flexible work schedules instead of having a dedicated desk for their exclusive use.
This approach directly benefits companies by increasing the rate of space utilization and reducing the cost of workspace \cite{mccoy2005linking}.
However, it presents the employees with a decision each time when they come to the office: choosing where to sit.

In desk sharing environments, workers must select a workspace that meets their needs for being in offices after evaluating necessary work-related information.
For example, workers may prefer a workspace with few other colleagues for focus work, while they may want to sit close to their collaborators for social connections or face-to-face discussion \cite{teevan2022microsoft, capossela2023people}.
To make an informed decision, workers may need more information than just basic data of workspace (e.g., location, availability), such as where their collaborators will sit today.
Furthermore, it’s not a straightforward task for employees to choose a desk given the multiple considerations and information they have: The employees must consider their schedules and work plans, navigate through the features of workspace, and then find one workspace that caters to their individual needs.
How workers select their workspace in desk sharing settings, including their considerations, priorities, and informational needs, has not been thoroughly explored yet.

The recent advancements in Large Language Models (LLMs) may help workers facilitate this process of selecting a workspace.
Several studies have shown that LLMs can perform complex tasks or make suggestions while taking into account much information \cite{liu2023using, ko2023can, rao2023evaluating, sorin2023large}. 
This capability may be leveraged to provide workspace suggestions tailored to the schedules and workspace preferences of employees.
Besides,  as indicated in many works \cite{umerenkov2023deciphering,martens2023tell, slack2023explaining}, LLMs have the potential to explain their suggestions, enabling the employees to understand how the suggestions are made before choosing whether or not to adopt the suggestions.

In this paper, we establish a decision support system for workspace selection by leveraging LLMs to provide workspace suggestions and accompanying explanations. 
We aim to examine the effectiveness, applicability, and user acceptance of the decision support system in the hybrid work environment. 
We set the following research questions to answer: 
\begin{enumerate}
  \item [(RQ1)] What factors do workers consider when selecting their workspace in a desk sharing environment? What factors are prioritized than the others? 
  \item [(RQ2)] How is the performance of LLMs in making workspace suggestions? What reasons do LLMs cite for their workspace suggestions?
  \item [(RQ3)] How do users of the LLM-empowered system perceive its effectiveness? Does the system facilitate their decision making process?
\end{enumerate}

In order to answer the research questions, we first conducted a pilot study (Section~\ref{sec:pilot}) with eight individuals to observe their typical daily work routine and the way they chose their workspace in the office, which informed how LLMs could assist in making workspace decisions. 
%The factors influencing their decisions can be categorized into those related to either physical (e.g., their assigned buildings, monitors) or social (e.g., collaborators, meeting attendees) environments.
Based on the results of the pilot study, we created a persona and a hybrid workplace scenario to evaluate LLMs' performance in workspace suggestions.   
In Section~\ref{sec:LLM_exp}, we analyzed 60 workspace suggestions and accompanying explanations generated by LLMs.
We focused on 1) workspaces suggested, 2) reasons provided, and 3) hallucination when analyzing the LLM outputs.
%Since the outputs of LLMs were probabilistic, we ran the LLMs for 60 times given the same synthetic desk sharing scenario to observe the patterns underlying the suggestions and explanations.
%For each run of workspace suggestion, the LLMs were prompted to propose three workspace options, each accompanied by three reasons.
%First, the top two suggestions were always the two workspaces familiar to the persona, on account of  regular location, collaborators and amenities.
%On the other hand, the third option served as an exploration for the persona to different workspace for purpose of focus work or networking.
%The LLMs also exhibited their understanding of space hierarchy and shared alternatives when some amenities were unavailable.
%Lastly, hallucinations were sometimes observed in the LLMs' explanations; the hallucinations did not distort the workspace information.
Finally, we conducted a user study in Section \ref{sec:user_study} to evaluate the effectiveness of LLMs in supporting the decision of workspace choices. The participants were asked to choose a workspace for a given scenario based on 1)  workspace information only, 2) additional workspace suggestions by LLMs, and 3) the reasons underlying the suggestions by LLMs. 

To the best of our knowledge, this is the first paper that experiments with LLMs for workspace suggestion systems. We summarize our main results and contributions as follows:
\begin{itemize}
    \item {\bf Reasoning ability and adaptability.} We investigated the capacity of LLM to suggest a suitable workspace based on the given information of individuals such as their meeting schedules and locations, coworker locations, and desk booking history. The LLM's reasoning extended beyond the guidelines provided in the prompt and could balance the trade-off among the given resources in the workspaces.
    \item {\bf Influence on user decisions.} The workspace suggestion system leveraging LLM had influence on user decisions. This effect could be greater when the benefits of the suggestions were explained to the users.
    \item {\bf User satisfaction and convenience of the system.} The respondents found the system to be convenient, regardless of whether reasons were provided or not. The system could reduce the users' effort for finding a suitable workspace. 
\end{itemize}

%When decisions are made by AI, users would like to know how such decisions are produced (ref). To provide good reasons for humans, it is required to understand them in terms of how they define, generate, process, and evaluate decisions and explanations (Miller, 2018; Datta and Dickerson, 2023). 

\section{Related Work}
\subsection{Physical and Social Factors in Workspace}

Interior office design affects workers' experiences in many aspects and several design strategies and measures have been identified to create work environments that meet the needs of workers \cite{arundell2018abw, debeen2014, engelen2019}. 
Most of all, workspace arrangement is a vital dimension in workplaces as it affects workers in various ways. 
Several papers examine how physical space affects the formation of social ties and discover that  proximity between workers increases the probability of communication and tie formation in a workplace \cite{allen1970, smallalder2019}.
The popular concept of desk sharing environments have been shown to improve communication \cite{de2005effect}, foster working relationships \cite{chigot2003controlled, mcelroy2010employee}, and promote the exchange of knowledge and skills \cite{ashkanasy2014understanding,chigot2003controlled}.
However, some research works indicate negative consequences of open architecture in offices: Workers tend to interact via emails rather than face-to-face in shared workspace \cite{bernstein2018impact}. 
It is also observed that desk sharing is linked to decreases in focus and increases in negative relationships, which may result from the distractions in a shared space and the hassle to find and accommodate to a space \cite{morrison2017demands}.
In this regard, our work focuses on what considerations workers may have for workspace selection and how an LLM-based system can effectively help workers navigate through the process, finding a suitable workspace for them.

\subsection{Large Language Models for Decision Making}

LLMs have been utilized to assist humans in a range of tasks that require text processing and generation \cite{lee2022evaluating, gero2022sparks, copilot, he2023survey,franceschelli2023creativity}. 
Moreover, several studies have proposed to use LLMs to handle complex problems that involve processing information and identifying conditions, and then make suggestions. 
For medical diagnosis, LLMs are leveraged as a support tool for breast tumor board \cite{sorin2023large}, radiologic decision-making  \cite{rao2023evaluating}, and clinical decisions \cite{liu2023using}.
In \cite{ko2023can}, it is shown that the performance of ChatGPT in portfolio construction is better than a portfolio with random selected assets. 
In \cite{wu2023exploring}, the authors  explored the LLMs' capability of parsing graph data and make job recommendations.
LLMs are also used to predict user ratings for items, given historical user interactions and preferences \cite{kang2023llms, zhiyuli2023bookgpt}.
It is also demonstrated that ChatGPT can be leveraged to enhance tourists' experience by providing customized recommendations  \cite{wong2023autonomous}.
In our study, we assess the effectiveness of LLMs to provide workspace  suggestions based on information related to a worker’s schedule and available workspace.

\subsection{Explainable AI and Large Language Models }

%However, if individuals do not trust the workspace suggestions provided by LLMs, they may not adopt them as their decisions. 
A recent line of research, known as explainable AI (XAI), suggests that users' trust in AI systems can be enhanced by improving their understanding of the AI’s decisions, including aspects such as transparency, traceability, and explainability \cite{doran2017does,adadi2018peeking,gunning2019xai,samek2019explainable, mueller2019explanation,lockey2021review}.
In the desk sharing environments, employees would like to know how workspace suggestions are generated and why they are beneficial.
%As many studies have shown \cite{umerenkov2023deciphering,martens2023tell, slack2023explaining}, LLMs can also play a role in offering explanations.

%AI technology has recently seen exponential growth and is now widely used in many aspects of our lives \cite{luckin2016intelligence,hamet2017artificial,imran2020using, pillai2020adoption,herath2022adoption}. 
%A survey conducted in \cite{Gillespie2023} over 17 countries shows that only 31\% of the respondents across the countries report that they are willing to trust AI systems.
%Previous research has demonstrated that trust is a significant determinant  of end users' willingness to adopt AI for various aspects of use \cite{komiak2006effects, qiu2009evaluating,  payne2018mobile, panagiotopoulos2018empirical, zhang2019roles}.

Several papers define what constitutes the explainability of an AI system \cite{lipton2018mythos} or outline the principles of XAI \cite{phillips2020four}.
In \cite{miller2019explanation} the author argues what constitutes effective explanations for people can be derived from the existing research in social science.
In particular, it is suggested in \cite{miller2019explanation} that people selectively care about only a few causes even though there can be numerous ones and that ``explanations should be presented as part of a conversation or interaction." 
Our approach is shaped by the insights: We initially conducted a pilot study aiming to identify the crucial factors in people’s process of workspace selection, based on which the system provided explanations. 

In the realm of explanation provision, LLMs have been used  to create narratives that elucidate the connection between system inputs and outputs, supplemented with human evaluations to assess the effectiveness of the LLMs' explanations \cite{umerenkov2023deciphering,martens2023tell, slack2023explaining}. 
In these studies, the outputs of the system can be either human-labeled data or predictions made by machine learning models other than LLMs.
By contrast, in our proposed system, LLMs are utilized not only to offer explanations but also \emph{to process the given information and suggest feasible options (i.e., output)}.
We further did an experiment on workspace suggestions and explanations generated by LLMs, scrutinizing their quality and attributes.

Similar to the systems that leverage XAI, the outputs of LLMs are not the final objectives for the humans and there is often no ground truth for the tasks they perform. Therefore, as suggested in \cite{datta2023s}, it is crucial to establish frameworks for human-centered qualitative evaluation about the impacts of LLMs' outputs on human objectives. 
%Furthermore, previous developments and advancements made by the XAI community can provide guidance in the study of LLMs. 
We conducted a user study to assess the effects of workspace suggestions by LLMs on employees' decisions and evaluate participants’ perception about our decision support system.

% We assess the effectiveness of the LLMs' explanations and also study whether the participants are satisfied with the suggestions made by the LLMs. We conducted a user study to evaluate the effects of workspace suggestions by LLMs on employees' decision on workspace selection. Subsequently, we evaluated participants’ perception on our decision support system when explanations supporting the workspace suggestions were provided.

\section{Pilot Study}\label{sec:pilot}
A pilot study was conducted in a technology company to explore work routines on a typical onsite workday and the decision-making process of workers when choosing where to sit in offices. 

\subsection{Methods}
The pilot study was carried out using a survey questionnaire that included two parts, summarized as below: 

\begin{itemize}
    \item The first section collected  background information of the surveyed participants, including their age, gender, job title, and work experience. Additionally, the participants were asked about their work styles and collaborations, such as the number of projects, the number of teams with which they were collaborating, and hybrid work patterns. 
    \item The second section focused on the participants' workspace choices, in the format of open-ended questions. The participants were asked to describe how they would select their workspace on a typical onsite day, and what information they would like to know in this decision-making process. They were also asked to indicate their preferred proximity between their workspace and the locations of their own collaborators, manager, or meeting venues. Finally, they were asked to share the types of information they would like to see in the explanations for  workspace suggestions. 
\end{itemize}

The responses to the open-ended questions were analyzed using two analysis methods. 
To find a typical onsite workday of the participants, we analyze the responses using the method suggested by Burnard  \cite{burnard1991method}: We used open coding methods to identify key behaviors as many as necessary and grouped them under high-order headings. On the other hand, to understand the decision process of the participants in workspace selection, we labeled  data by conducting initial coding as a first approach of a grounded theory coding method.

\subsection{Participants}

The eligibility for participation in this study was as follows: (1) US-based, (2) a minimum age of 18 years, and (3) employment in an onsite or hybrid work environment (not fully remote).
Eight workers were recruited for our pilot study. 
They were informed of the study’s objectives, provided their consent, and were made aware of the potential benefits of the survey.
As a token of appreciation for their time and effort, they each received a USD \$15 gift card.

The job titles of the participants included researcher, data scientist, software engineer, and product designer. Half of the participants were female.
62.5\% of the participants had at least 1 year of work experience, while the remaining participants had more than 10 years of experience.
The number of projects they were working on ranged from 2 to 4. One participant worked 100\% in-person (5 times a week), and the others worked in hybrid mode.

\subsection{Pilot Study Results}\label{sec:pilot_result}

\indent {\bf  A typical onsite day of the participants.}
The participants indicated they typically had 2 meetings. Meetings were mostly virtual but could be in person or hybrid if other attendees were in offices. The other work time was usually split between focus work and team collaboration.

{\bf Decision factors considered in workspace selection.}  We found that the factors could be categorized into two aspects of work environments: \emph{physical and social environments}. First, all participants indicated physical attributes of the workspace as crucial factors in their decision-making process. Specifically, the attributes included (1) their assigned buildings, (2) proximity to meeting rooms, (3) privacy, (4) availability of devices (e.g., monitors), and (5) a view.
On the other hand, the participants mentioned that their choice of workspace was also influenced by social factors, particularly the proximity to their key collaborators. These collaborators typically included (1) individuals working on the same project, (2) their manager, and (3) individuals scheduled for meetings on the workday. The participants also suggested that the seniority of their collaborators played a role in their decision to sit close to them.

However, the participants had diverse preferences on the level of proximity to their key collaborators or meeting rooms. Five out of the eight participants preferred to stay ``in the same neighborhood\footnote{A neighborhood is defined as an open office area that accommodates between 6 to 20 workstations, facilitating close interaction among employees.} '' with their collaborators, while the others defined ``on the same floor'', ``in the same building'', ``on the same campus'' as a close location, respectively. On the other hand, when they considered close locations to meeting rooms, ``in the same neighborhood'' or ``on the same floor'' was defined as a close location by one participant. 

Lastly, some participants mentioned that not all information for their decision making was easily accessible, such as occupancy status, equipment (e.g., whiteboards, monitors), and collaborator locations. 

\section{Workspace Suggestions by LLMs}\label{sec:LLM_exp}
In order to validate the applicability of LLMs for workplace decision support system, we used LLMs to generate workspace suggestions and reasons, and examined the quality of the output. 

The experiment was based on a synthetic persona that was created using the work-related information collected in the pilot study, as described in Section~\ref{sec:pilot_result}. The persona had 2 in-person meetings, 3 key collaborators, and a manager on campus on their onsite work day. A total of 6 neighborhoods were available for selection (4 and 2 neighborhoods in two different buildings respectively). 

\subsection{Prompt and Settings}

To improve the quality of reasoning by LLMs, we adopted a few-shot learning strategy in which the LLMs were provided with several examples of reasoning as guidance for its output. For example, we wrote ``This workspace is on the same floor as your meeting room for [meeting subject]'', and ``Your teammate, [teammate name], is also sitting here today, so you can collaborate with them easily.'' 

The workspace suggestions were generated using GPT-4.
The parameters were configured as follows: temperature = 0.1, maximum token = 700, and top\_p = 1.
The prompt for the LLMs consisted of 6 parts: (1) goals and tasks, (2) explanation for building information, including location hierarchy (neighborhood -- floor -- building), (3) data description and data format, (4) the types of workspace and their features, (5) examples of reason, and (6) data input.
In particular, in the (1) goals and tasks, the LLM was instructed to output its top 3 neighborhood suggestions out of 6 available workspaces and provide 3 reasons for each workspace suggestion in bullet points, given the persona and available information. 
The synthetic data and prompt are attached in Appendix. 

To examine the performance and consistency of the LLM in workspace suggestion, we executed 60 runs of the workspace suggestion given the scenario (We will refer to the LLM's output in the $n$th run as Output $n$.).

\subsection{Workspace Suggestions and Accompanying Reasons Provided by LLMs} \label{sec:llm_experiment}
In each run of workspace suggestion, the LLM provided its top 3 workspaces that might cater to the persona's need and provided accompanying explanations.
The summary statistics of the suggestions with respect to their rankings are presented in Table \ref{tab:frequency}.
Five out of 6 workspaces were ever suggested by the LLM. Neighborhood 2402 and 2403 in Building A appeared in all of the outputs in either the first or second rank.
For the third rank, neighborhood 2504 in Building A was suggested most frequently, followed by Building B 4309 and 4215. 
Below are our observations on the suggestions generated by the LLM.

\begin{table}
    \caption{Frequency of workspaces suggested by the LLM}
    \label{tab:frequency}
    \begin{tabular}{ccccc}
        \toprule
         Workspace & rank 1& rank 2&  rank 3& Total\\
         \midrule
         Building A 2402&  48& 12 & 0 & 60\\
         Building A 2403&  12& 48 & 0 & 60\\
         Building A 2504&  0&  0& 55 & 55\\
         Building B 4215&  0&  0& 1 & 1\\
         Building B 4309&  0&  0& 4 & 4\\
         \bottomrule
    \end{tabular}
\end{table}

{\bf Types of considerations.} 
We categorized the LLM's considerations in 60 outputs into six types: 1) regular workspace location, 2) collaborator location, 3) amenity (i.e., monitor availability and wheelchair accessible), 4) meeting room location, 5) availability over capacity (crowdedness), and 6) non-regular workspace. 
It is worth noting that though it was instructed in the prompt that the system could consider desk-booking history for the LLM's suggestions, the LLM never mentioned it as consideration.

The LLM explained its considerations with purposes or specific information. 
For regular workspace location, the LLM mentioned ``\emph{you can easily access your meetings and amenities (or collaborators)}" without specific details, which referred to familiarity of the working environment.
For considerations (2)-(4), the purposes were self-evident and the LLM offered detailed information from input data.
When considering availability (crowdedness), the LLM only proposed workspaces with low occupancy rate and explained it could be used \emph{for focus work}. 
Lastly, when the LLM pointed out a workspace was ``\emph{not your regular workspace}," it explained that the workspace might serve for networking purpose (e.g., ''\emph{so you can network with other colleagues and learn from different perspective [Output 12]}") or for focus work (e.g., ''\emph{might have more privacy and less distraction [Output 25]}).

{\bf Frequency and priority of considerations.}
Table~\ref{tab:OutputFreq} shows the frequency of the confederations appearing in the top 3 workspace suggestions. First, we found that regular location and amenities were considered frequently in all ranks. Next, collaborator location was prioritized in the first and second ranks, and meeting room location was often cited as a reason in the second suggestion.
Lastly, availability and non-regular workspace appeared more frequently in the third suggestion, indicating the LLM's emphasis on focus work or networking opportunities.
We also note that for each suggestion the LLM often mentioned more considerations than instructed in the prompt (i.e., 3 considerations); this was reflected in the total of considerations in each rank (which should have been 180 if following the prompt).

\begin{table}
    \caption{Type and frequency of considerations in the 60 outputs of workspace suggestions by the LLM} 
    \centering
    \begin{tabular*}{\linewidth}{@{\extracolsep{\fill}}cccc}
         \toprule
         Consideration             & rank 1 & rank 2 & rank 3\\
         \midrule
         Regular location*        & 57      &  58      & 53   \\
         Collaborator location     &  58      &  60      & 7    \\
         Amenity                    &  44      &  47      & 58   \\
         Meeting room location      &  16      &  51      & 39   \\
         Availability (Crowdedness) &  16      &  12      & 60   \\
         Non-regular workspace      & 0        & 0        & 26  \\
         Total                      &  191     &  228     & 266  \\
         \bottomrule
    \end{tabular*}
    \begin{tablenotes}[para,flushleft]
      \small
      \item Note: Regular location includes regular workspace, floor, and building. 
    \end{tablenotes}
    \label{tab:OutputFreq}
\end{table}

{\bf Understanding the hierarchy of spaces.} The LLM could distinguish the hierarchy of the spaces (e.g., regular workspace, regular building and floor). 
For example, in Output 43, \emph{``this workspace is where you usually sit"} referred to the persona's regular workspace A2402.
On the other hand, in Output 2, the LLM suggested A2504, which was not the regular workspace, with 2 reasons stating \emph{''this workspace is in your regular building and floor, and close to your meeting room (Conf Room Building A/2011) for the Data science research review; this workspace is in a different neighborhood from your regular one (workspace)."}
This showed that the LLM could tell workspace A2504 from A2402, and identified that they shared the same building and floor where the persona regularly visited.

{\bf Alternatives for amenities.} The LLM would offer alternative solutions if some amenities were not available in its suggestion. For example, in Output 52, it explained for workspace A2504, \emph{``The workspace does not have a monitor available. You can bring your own laptop or use a nearby focus room if you need a monitor."}

%\ch{I think we can remove it.}{\bf Environments for focus work.} When the occupancy rate of a workspace is low, it is regarded as suitable for focus work. We observed that GPT-4 explained for its suggestion workspace A2504 in Output 25, \emph{``This workspace has 8 capacity, and only 1 seat is occupied, so you can have plenty of space for focus work."} Moreover, GPT-4 also pointed out another reason for focus purpose. In the same suggestion, GPT-4 added that \emph{``this workspace is not close to any of your collaborators or teammates, so you might have more privacy and less distraction."}

\begin{table*}[ht]
    \caption{Examples of the LLM's hallucination in workspace suggestions} 
    \centering
    \begin{tabular}{ccc} \hline
    \toprule
    Type & Output example & Problem description\\ \midrule
    Illogical reasoning  & \multicolumn{1}{p{8cm}}{ This workspace does not have a monitor, so you can \emph{avoid distractions from other screens}. [Output 38] }& \multicolumn{1}{p{5.5cm}}{This explanation was not reasonable.} \\ 
    \hline
    Information fabrication   &   \multicolumn{1}{p{8cm}}{You have booked this workspace \emph{four times in the past 15 days}, so you might be familiar... with the location. [Output 20]} & \multicolumn{1}{p{5.5cm}}{This frequency of desk booking history was not given.}    \\
         &  \multicolumn{1}{p{8cm}}{This workspace is \emph{wheelchair accessible, which can accommodate your needs}. [Output 32]} & \multicolumn{1}{p{5.5cm}}{The persona didn't specify any requirement. } \\ 
         &\multicolumn{1}{p{8cm}}{This workspace (A2504) is in a \emph{high-traffic area}, so you can network...learn from different perspectives. [Output 33]} & \multicolumn{1}{p{5.5cm}}{There is no information about traffic around workspaces. Besides, the workspace had a low occupancy (one out of eight desks was occupied). }    \\ 
         &  \multicolumn{1}{p{8cm}}{You can bring your own laptop or use \emph{a nearby focus room} if you need a monitor. [Output 52]} & \multicolumn{1}{p{5.5cm}}{Workspace information other than neighborhoods was not provided. }  \\ 
    \hline 
    Unconfirmed speculation   &  \multicolumn{1}{p{8cm}}{Your collaborator Marcus Fields is also sitting here today, so you can work with him on \emph{your data science research project}. [Output 29]}   & \multicolumn{1}{p{5.5cm}}{This research project was speculated from the meeting information but such a project was not explicitly specified.}\\
         
    \bottomrule
    \end{tabular}
    \label{tab:hallucination_wronginfo}
\end{table*}

{\bf Hallucination.} We observed hallucinations in the explanations by the LLM and classified them into three types as shown in Table \ref{tab:hallucination_wronginfo}. The first type of hallucination  was \emph{illogical reasoning} for which accompanying explanations were not logically reasonable. Second, the LLM \emph{generated information} that was not in the data section of the prompt, e.g., frequency in the persona's booking history, the persona's requirement, and building information other than workspaces. 
Lastly, LLM made some inference that was considered \emph{speculative or unconfirmed}.
We remarked that none of the hallucinations distorted the information and data we input.

% \begin{itemize}
%     \item [Output 20] You have booked this workspace \emph{four times in the past 15 days}, so you might be familiar...with the location. -- This frequency was not given. 
%     \item [Output 32] This workspace is \emph{wheelchair accessible, which can accommodate your needs}.  -- The persona didn't specify this requirement.
%     \item [Output 33] This workspace (2504) is in a \emph{high-traffic area}, so you can network...learn from different perspectives. -- The workspace 2504 had a low occupancy rate (1 occupied out of 8).
%     \item [Output 52] This workspace...You can bring your own laptop or use \emph{a nearby focus room} if you need a monitor. -- There was no such information.
% \end{itemize}

% \begin{itemize}
% \item [Output 29] Your collaborator Marcus Fields is also sitting here today, so you can work with him on \emph{your data science research project}. -- GPT-4 inferred this research project from the meeting information but we did not specify this project. 
% \item [Output 38] This workspace does not have a monitor, so you can \emph{avoid distractions from other screens}. -- This explanation was not reasonable.
% \end{itemize}

\section{User Study} \label{sec:user_study}
The primary goal of the user study is to evaluate the effectiveness of the LLM-empowered workspace decision support system. 
In this user study, participants were asked to choose a workspace based on the given scenario, with and without the suggestions produced by the LLM. We also examined how participants perceived the effectiveness of making decisions on the workspace with the suggestions provided by LLMs. 
The scenario in the survey was the same synthetic persona that was used in the LLM experiment, described in Section~\ref{sec:LLM_exp}.
%The scenario was developed based on the information collected about the participants’ work routines in our pilot study. 

\subsection{Survey Questionnaire}
The questionnaire consisted of two sections. The first section included eligibility questions for the survey participation and asked for participants' consent. The respondent were also asked to share their personal experiences of workspace use, such as their current workspace settings, experience in sharing desks, and hybrid work patterns.

The second section consisted of three phases in which the participants were required to make decisions about where to sit if positioned in a given scenario. 

%The scenario was developed based on the information collected about the participants’ work routines in our pilot study. 
Under the given scenario, phase 1 required participants to decide where they would like to sit, without any suggestions provided.
On the other hand, in phases 2 and 3, the participants were asked to choose their workspace based on the same scenario and given \emph{a list of three workspace suggested by LLMs}. The list was chosen from our experiment in Section \ref{sec:llm_experiment}: We included two workspaces that were always suggested (i.e., Building A Neighborhood 2402 and A 2403) and one workspace from a different building (i.e., B 4309) for diversifying workspace features.
However, in phase 2 only the suggested workspaces were provided, while in phase 3, each suggestion was accompanied by \emph{three reasons} explaining why the corresponding workspace was recommended. 

In the phases 1-3, the participants were also asked to evaluate the decision-making process. 
Specifically, in the phases 1-3, they were required to choose their level of confidence and convenience about their choices at a 5-point Likert scale (1: Not at all to 5: Extremely). 
Additionally, in the phases 2 and 3, they were asked to express their level of satisfaction with the suggestions by LLMs using a 5-point Likert scale (1: Not satisfied at all to 5: Extremely satisfied). 
Lastly, in the phase 3, the participants were asked to evaluate the level of relevance of the reasons in a 5-point Likert scale (1: Not relevant at all to 5: Extremely relevant) and to provide what other reasons they would like to see while making a decision in open-ended question. 

\subsection{Participants}
The survey data were collected in a global technology company. The recruitment was posted on an internal corporate social network and sent out via email. The eligibility for the main study included (1) US-based, (2) at least 18 years old, (3) not entirely working remotely, and (4) not in sensitive roles. 
A total of 49 responses were collected, but only 42 of them were valid responses and were used for analysis. The job titles of respondents varied, including 13 manager, 11 software engineers, 6 scientist or researchers, 5 Product managers, and 7 others. Two respondents did not have any assigned building nor workspaces, and the others (40 respondents) were assigned to a building or a workspace (see Table \ref{tab:hybridwork} for details). 57\% participants had experiences with desk sharing. For hybrid work patterns, two respondents fully worked in their offices, while the others came to offices on average 1-2 times (59.5\%) or 3-4 times (35.7\%) every week.
We will refer to the $n$-th respondent entering our survey by $Rn$.

\begin{table}
    \caption{Workspace settings of the respondents}
    \centering
    \begin{tabular}{cc}
         \toprule
         Workspace assignment type  &  Count\\
         \midrule
         Assigned desk in a private office (capacity 1-2) &  8   \\
         Assigned desk in an open office   &  9    \\
         Assigned neighborhood but no assigned desk   &  12      \\
         Assigned building but not assigned workspace &  11     \\
         No assigned building &  2      \\
         Total  &  42     \\
         \bottomrule
    \end{tabular}
    \label{tab:hybridwork}
\end{table}

\subsection{Analysis}
We first investigated the importance of decision factors in selecting workspaces, by checking their frequency in the responses. After that, we tracked the changes in the workspace choices of the respondents over the three phases in our survey. 

To analyze how people perceived the effectiveness of LLMs over decision phases, we employed within-subjects ANOVA (one-way repeated measure ANOVA) to identify the main effects of decision phases and random effects of individuals. The dependent variables for ANOVA were the perceived confidence and convenience when they made decisions. Additionally, post hoc analysis was performed using a Bonferroni method. Lastly, we performed a paired t-test to test the differences in the perceived satisfaction with the suggestion between the second and the third phases. 

\subsection{Results}

\subsubsection{Considerations for selecting workspace}
\begin{figure}[t]
  \centering
  \includegraphics[width=\linewidth]{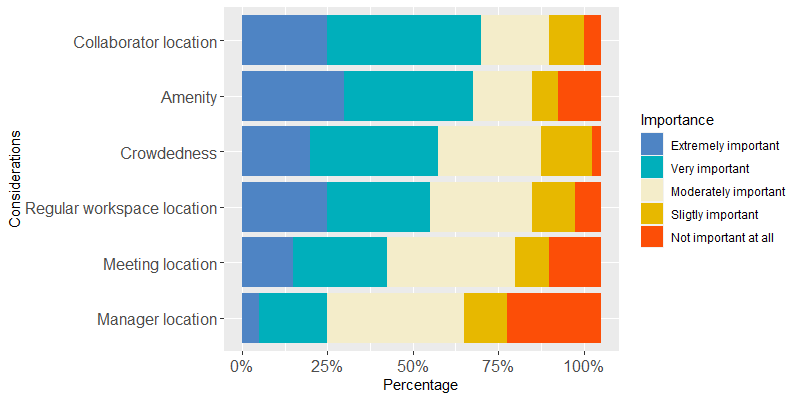}
  \caption{The survey results of the importance of considerations for deciding workspace location. The considerations are displayed from the highest to lowest average by converting responses on a Likert scale into numeric values (1: Not important at all to 5: Extremely important): collaborator location (3.714), amenity (3.619), crowdedness (3.548), regular workspace location (3.500), meeting room location (3.167), and manager's location (2.643).}
  \label{fig:con_freq}
\end{figure}

Figure \ref{fig:con_freq} presents the considerations in an order of their importance in the respondents' selection for a workspace. The location of collaborators was the most important consideration as over 67\% of respondents scored it extremely  or very important. Next, 64\% of respondents regarded amenity as an important consideration. The considerations in the subsequent ranking were crowdedness, regular workspace location, and meeting location, rated as important by 55\%, 52\%, and 40\% of the respondents, respectively. Lastly, the least important consideration was manager location with only 23\% of the respondents thinking it as important.

\subsubsection{Users' workspace choices using the decision support system}
%How useful can the LLM-leveraging workspace suggestion system be? 
We examined how the respondents' choices of workspace might change with the suggestions and explanations by the LLM. 

In the 1st phase, five of the six workspaces were ever chosen by the respondents:  A2402 (23 respondents), A2403 (16), A1401 (1), A2504 (1), or B4309 (1). The distribution of the choices concentrated on A2402 and A2403, showing a similar pattern to the one in the first-rank suggestions by the LLM (A2402 (48/60 outputs) and A2403 (12/60 outputs)). 
In the 2nd phase when the suggestions by the LLM were provided, 38 respondents out of 42 still selected the workspace same as their choice in the 1st phase. 
All of the respondents' choices were regressed to one of the suggestions by the LLM, i.e., either A2402, A2403, or B4309.

In the 3rd phase when the explanations were also provided, even though all of their choices were still from the LLM's suggestions, 10 respondents changed their decisions from their choice in the 2nd phase. 
Some of these respondents commented that they changed their choices considering the reasons provided. For example, when the reasons cited the consideration of the respondents, they would follow the suggestions with those reasons, e.g., R20 stated \emph{``...there are monitors there I would head there.''}
Another case was that the respondents found the reasons acceptable and were willing to take the suggestion, e.g., \emph{``Suggestions made it clear that it is convenient here [R15].''} 

On the other hand, some of the respondents who did not change their choices in the 3rd phase explained that the system offered acceptable reasons, e.g., \emph{``with the added benefit that neighborhood 2403 is on the same floor as my meetings [R18].''} 
One of the respondents even mentioned that \emph{``I like the logic - the benefits outweigh the others listed in other options - and the fact that there's nobody else working there.''}

\begin{figure}[t]
  \centering \includegraphics[width=\linewidth]{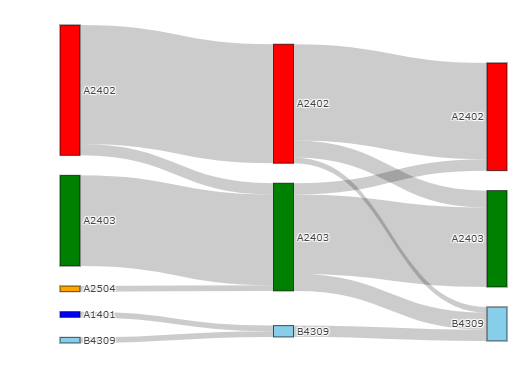}
  \caption{The workplace choices over selection phases. Each level represents decision phase, showing the flows of choices from the first phase to second and third phases. }
  \label{fig:choice}
\end{figure}

\subsubsection{User evaluation on the decision support system}
\begin{figure}[t]
  \centering
  \includegraphics[width=\linewidth]{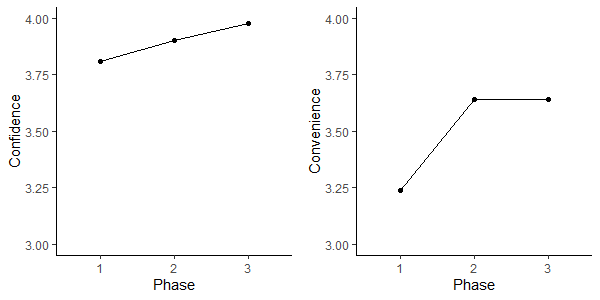}
  \caption{The level of perceived confidence and convenience in decision process. Phase 1; decision by user. Phase 2; workspace suggestion by LLM. Phase 3; workspace suggestion with reasons by LLM.}
  \label{fig:plot_repeated}
\end{figure}

In each phase of our survey, the respondents were required to express their perceived level of confidence and convenience about their decision making after they made a decision given the information (and suggestions). 
Figure~\ref{fig:plot_repeated} shows the average level of confidence and convenience in each phase.
We found that the average of the perceived confidence about their decisions increased from 3.81 to 3.90 to 3.98 over the three phases but the ANOVA model was not statistically significant $(F = 1.795)$. 
However, the perceived convenience changed significantly $(F = 7.6, p<.005)$ over the phases, which was 3.24 in the first phase and 3.64 in the second phase and the third. The increases were significant from the first to second $(t = -3.42, p<.005)$ and third $(t = -3.29, p<.005)$ phases, whereas there was no significant differences between the second and the third phases. 
%and. The convenience in the second and third phases was significantly higher than that in the first phase, but there was no significant differences in the perceived convenience between the second and third phases. 

In the second and third phases, the respondents also indicated their level of satisfaction with the suggestions. The average level was improved slightly from 3.74 (without explanations) to 3.79 (with explanations), showing no significant difference $(F = 0.06)$. 

The respondents were asked for their opinions about the number of suggestions. Most of them were satisfied with 3 suggestions as in the survey. However, five respondents would like to see a different number of suggestions: two of them desired for 5 suggestions; one indicated between 8 and 10 suggestions; one said two suggestions. The other respondent did not specify their desired number of suggestions.

\section{Discussion}
In this section, we discuss the experiment results of the LLMs generating workspace suggestions and the implications of our user study. We will highlight several aspects of the system that remain unexplored in this study, presenting them as potential avenues for future research.

\subsection{Leveraging LLMs for a Decision Support System for Hybrid Workspace} \label{sec:discussion_effectiveness}
Reasoning ability is one of the significant advantages of LLMs and has been tested in many studies. In this study, we explored utilization of an LLM for a decision support system, examining its capabilities of generating suggestions and accompanying reasons.

First, the reasoning of LLMs met workers' considerations and satisfied their informational needs. 
In the prompt, we presented a list of factors for LLMs to consider; we did not assign any ranking of importance to them. 
It turned out that the factors the LLM prioritized were similar to those the participants found important. 
Figure \ref{fig:con_freq} shows that individuals identified collaborator location, amenity, and regular workspace location as important factors. These factors were also frequently referenced by the LLM, as depicted in Table \ref{tab:OutputFreq}. 
On the other hand, when the information in the reasons by the LLM was limited, people sometimes relied sorely on the reasons given by the LLM, without checking available information. This behavior underscores the potential of employing LLM as a decision support system for workspace, presenting that LLMs "naturally" do the reasoning in this task well.

Second, we noted that the LLM’s reasoning extended beyond the guidelines provided in our prompt and could be adaptable for different workspaces.  %We provided some examples of reasoning in the prompts, and the LLM provided other reasons that were not given in the prompt. 
For example, the LLM mentioned non-regular workspace as a reason for workspace suggestion. Interestingly, in addition to the facts, LLM was able to provide potential benefits such as focus availability and potential to socialization. Another notable point is that LLM provided alternatives under a given situation. That means LLM can balance the trade-off among the given resources in the workspace in decision making. 
%By using LLM, workers can utilize workspaces in not only various but also helpful ways to their work. 

Third, in the LLM's workspace suggestion we identified an \emph{explore-exploit strategy}. 
The so-called explore-exploit behavior is often observed in processes of human decision making.
This involves a choice between exploitation, by which decision maker pursues a known benefit, and exploration, by which they sample options with uncertain outcomes in search of potential gains \cite{wilson2017exploit}. 
In the context of workspace selection, options for \emph{exploitation} are the workspaces with which workers are familiar, especially their regular workspace; they know well the advantages and disadvantages of using these workspaces.
On the other hand, options for \emph{exploration} refer to the non-regular workspaces which the workers have rarely or never used. 
We can see in Table~\ref{tab:frequency} and \ref{tab:OutputFreq} that the first two suggestions were options for exploitation whereas for the third rank the LLM often pointed out that the workspace was non-regular and cited potential benefits. 
This explore-exploit strategy in the decision support system allows for the fact that workers do not always have the same intent of coming to the office. 
Their considerations can change according to their work plans and offering options for exploration can be beneficial to satisfy different considerations than regular routines. 
The explore-exploit strategy employed by LLM in the decision support system is able to offer a range of decision options.
If employees frequently have varying considerations, it may be beneficial to explicitly direct LLMs to use this strategy.

This decision support system also showed robustness and reliability. 
Since the system leverages LLMs, it is subject to the randomness and hallucination that are inherent in LLMs. 
However, in Section \ref{sec:llm_experiment}, we observed little variation in the LLM's output of workspace suggestions, where for each rank there were only 2 or 3 possible outcomes and the most frequent outcome accounted for at least 80\% of the outputs.\footnote{Precisely, 80\% for A2402 in rank 1, 80\% for A2403 in rank 2, 91.7\% for A2504 in rank 3.} 
The hallucinations in explanations were infrequent: For any workspace suggestion, at most one of the reasons was hallucinated. Additionally, these hallucinations did not distort any information and data we provided, posing little harm on the suggestions themselves.
%, where  SpecifLLM was able to provide consistent suggestions, showing 80\% outputs suggested Building A Neighborhood 2402 as the first rank. 

\subsection{Effects of LLM-empowered Workspace Suggestion System on a User's Decision}
The workspace suggestion system empowered by LLMs can hold the potential to influence user decisions. Despite the lack of a substantial increase in their perceived confidence, certain respondents of this study found values in LLM’s workspace suggestions, modifying their decisions. 
Especially, when explanations were provided with the benefits of each suggestion highlighted, the system had a greater effect on users' decisions: The respondents could readily identify the workspace that met their individual needs.
It is beneficial to users because in hot-desk settings satisfying personal preferences is crucial for work productivity \cite{kim2016deskownership}, and LLMs can facilitate this searching process with its reasoning capability. Moreover, the LLM's ability to provide reasons for its own suggestions fosters transparency, allowing users to understand the LLM's suggestions and make informed decisions.

Indeed, the positive responses collected from our survey support the use of the system that utilizes LLMs for decision-making assistance.
The respondents found the system to be convenient for workspace selection, regardless of whether reasons were provided or not. The system could cater to users’ needs and preferences, reducing their time and effort for finding a suitable workspace. 
Overall, the workspace system, empowered by LLM,  was well-received by the participants and proved effective in aiding their decision-making process, thereby showcasing its potential to improve work experience and boost employee satisfaction.

\subsection{Extensions and Limitations}

Below we discuss several aspects that were not covered in the system, which can be potential extensions of our work.

{\bf Visual aids.} Visual aids can enhance the process of selecting a workspace by providing a clear representation of workspaces, meeting rooms, and seating arrangements.
In our study, we presented the information in text format, which required users to read and interpret the data. 
One of the respondents stated \emph{``I'd like a picture attached to each option since the work environment matters too [R4]. ''}
Incorporating visual elements such as photographs of the workspace and floor plan layouts could streamline this process and make information processing more efficient, facilitating the decision making.

{\bf Personalized experience.} We personalized the user experience in workspace suggestions by incorporating personal information such as previously chosen workspaces, meeting details, and the locations of collaborators. 
One way to enhance the user experience is by customizing workspace considerations for each weekday, taking into account the regular routines of workers throughout the work week.
Specifically, a worker may tend to use Fridays for focus work and flexible time with less distraction from coworkers, and this user behavior can be captured in the system with re-arranged rankings of workspace suggestions.

{\bf Open-ended preferences.} We may enable the system to  capture nuanced personal preference or special needs by including \emph{chat interaction leveraging LLMs}.
For example, a participant said temperature of workspace was a main concern as they always felt cold, which was a factor not on our list in the prompt.
This case might be addressed via a chat interaction between the participant and LLMs during which the participant further clarified their input (e.g., a preferred range of degrees) and the LLMs identified their need and made suggestions.
The implementation of chat experience requires that the system allow LLMs to retrieve a wider range of information about workspace, employees, and work contexts.

{\bf Communication and coordination.} In this paper, we investigated how a worker's decision making for workspace could be improved with the help of LLMs \emph{given other users' decision (where they would sit) at the timing of using the system}. 
However, the intended plan of this worker may be impacted by a subsequent workspace decision made by another worker. 
For example, Alice first chooses a workspace with a low occupancy rate to fully focus on her paperwork today. 
Her collaborator, Bob, without knowing her intent, later selects the same workspace planning to discuss with Alice about their project. 
The Bob's decision may then disrupt Alice's plan today and her productivity. 
This suggests that the procedure of choosing a workspace should necessitate dialogue and collaboration among colleagues, ensuring their objectives are recognized and accommodated.
This communication channel is absent in our system. 

{\bf Privacy.} The aforementioned functions involve much diverse information as foundation support.
However, except for the public accessible information such as workspace or busy-idle employee calendars, employees may be uncomfortable with sharing more personal information for system use and coordination due to privacy concern. This requires further user studies to understand the balance between improvements on user experience and their privacy.

\section{Conclusion}
Our study demonstrated the capacity and explainability of LLMs in the development of a workspace decision support system, particularly in providing reasoned suggestions. 
% We first reveal the capacity and explainability of LLM as a decision support system for workspace in the hybrid work era. 
We identified the key considerations of workers when choosing where to sit in offices.
Our experiment revealed that LLMs were capable of recommending suitable workspaces based on relevant information and explaining its rationale beyond the suggestions. 
% We recognized the importance of providing explanation during the process between interaction between the system and humans from the XAI perspective. 
Furthermore, we explored how the explanations supported individuals in their decision making process of selecting a workspace. 
We observed that the employees found their ideal workspace more conveniently with the assistance of the LLM-empowered system.
Our results contribute to research on XAI and decision support systems for workspaces in hybrid era, and complement the recent literature on using LLMs for handling complex decision making.

\section{Ethics Statement}
This research includes user studies that may raise potential privacy concerns of the participants. To address this concern, our organization’s legal team conducted a formal privacy review and approved this study, ensuring that the participants’ identities remain confidential and protected. The participation was voluntary, and all participants were provided with a consent form that explained the objectives and process of the study, potential risks, and their rights. The data collected during the study was anonymized and analyzed at a higher-level aggregation to protect the privacy of the participants. 

\bibliographystyle{ACM-Reference-Format}
\bibliography{References.bib}

\appendix

\section{Scenario of Workspace Decision }
Imagine you are Yasmin and aim to find a workspace for August 8, 2023, considering the given information below.

{\bf Meetings:}
\begin{enumerate}
    \item Data science research review 
- Time: 2:35PM - 3:30 PM
- Location: Building A/ 2011
- Attendees: Marcus Fields (organizer), Yasmin Marsh (you), Jakobe Cain, Glenn Maxwell

\item 1:1 weekly sync Amanda | Yasmin
- Time: 4:05 PM - 4:30 PM
- Location: Building A/ 2016
- Attendees: Yasmin Marsh (You, organizer), Amanda Dalton (your manager)
\end{enumerate}

{\bf Workspace information on August 8, 2023:}
\begin{enumerate}
\item Building A Neighborhood 1401
- Jakobe Cain is sitting here.
- Features: Monitor
- Availability: 4/6

\item Building A Neighborhood 2402
- Your regular workspace
- Your manager (Amanda Dalton) is sitting here
- Features: Monitor, wheelchair accessible
- Availability: 3/8

\item Building A Neighborhood 2403
- You have used this workspace before
- Your collaborator, Marcus Fields, is sitting here
- Features: Monitor, wheelchair accessible
- Availability: 7/10

\item Building A Neighborhood 2504
- Features: Wheelchair accessible
- Availability: 7/8

\item Building B Neighborhood 4215
- Your collaborator, Glenn Maxwell, sits here
- Features: Wheelchair accessible
- Availability: 5/10

\item Building B Neighborhood 4309
- Features: Monitor, wheelchair accessible
- Availability: 8/8 
\end{enumerate}

\section{Prompt for Workspace Suggestion System}
You are an advanced AI assistant for workspace suggestion. Your job is to find the best desk where fits to the employee's employee of the day when they come to office.
You have access to the employee information, meeting locations, teammates, collaborators, desk booking history, and workplace information.
Important things to consider when suggesting a desk location
\begin{itemize}
    \item building the employee regularly sit in
    \item where collaborators/ favorites are sitting
    \item where the employee's team is sitting
    \item meeting locations
    \item desk booking history.
\end{itemize}

Note for building information and location hierarchy:
\begin{itemize}
    \item The number of available seats is not important if it has at least one.  
    \item Location hierarchy: Building - floor - neighborhood. 
    \item The rooms located in different buildings are not close to each other.
    \item You can consider proximity between rooms and desks only based on coordinate and location hierarchy.
    \item Keep in mind that floor 4 is different to floor 5.
    \item You might not get all data you need. You will need to provide the best suggestions based on the given data. 
    \item You do not know the distance between buildings.
    \item The information about wheelchair accessibility is given in location data ("IsWheelchairAccessible":true). 
    \item You do not know the locations of restrooms and elevators unless it is given. You never mention their locations if you do not have concrete evidence.
    \item If there is no available workspace that meets the employee's needs, you can explain it and provide alternatives.
\end{itemize}

Note for data use:
\begin{itemize}
    \item The information that you can use is provided after chevrons with the data name.
    \item You must provide outputs based on the given information and data (worker and space information).
\end{itemize}

Note for workspace type:
\begin{itemize}
    \item The conditions for focus work are the low density of people (fewer number of people in the same space). If the neighborhood is occupied over 60\%, the space might not be good for focus work.
    \item You always find desk in private office or neighborhood. Employees do not work in meeting rooms. If the employee wants a desk close to meeting rooms, find one in either private office or neighborhood. 
    \item Neighborhood: A neighborhood workspace refers to an open-space area with typically 6-20 desks. This workspace type encourages collaboration and communication with one another within the same neighborhood. This might not be the best for focus work, but people can find focus rooms or meeting rooms near their neighborhood.
    \item Private office: A private office has capacity of one or two. This office type is similar to focus room and preferrable for focus work or online meetings.
    \item When meeting rooms are close to the workspace, people can save time to travel. 
\end{itemize}

Note for reasons
\begin{itemize}
    \item Provide details of your reasoning  if possible (e.g., collaborator names, teammate name, meeting location (building and floor), meeting titles).
    \item The employee knows where their regular building is. Do not provide this for a reason.
    \item Be creative to provide suggestions and reasons but should be based on a given data and information.
    \item Personalized reasons are preferred. 
    \item Try to provide the different reasons for each suggestion. 
\end{itemize}

Examples of reasons:
\begin{itemize}
    \item This workspace is on the same floor as your meeting room (Building D 2601) for the team All Hands.
    \item This workspace has 9 capacity, and no one is sitting here today, so you can have plenty of space for focus work.
    \item Your teammate Amber Parrish is also sitting here today, so you can collaborate with them easily.
    \item This workspace is in a high-traffic area, so you can network with other colleagues.
    \item You never use these examples to provide output to the employee. You can modify and use the example reasons.
    \item You can create reasons. A variety of reasons for each suggestion is desired.
\end{itemize}

Note for output:
\begin{itemize}
    \item Provide maximum 3 suggestions. 3 suggestions are desired.
    \item Provide maximum 3 reason in bullet points.
    \item You can refer the employee "you" when you provide outputs.
\end{itemize}

\end{document}